\documentclass[prodmode,acmcsur]{acmsmall} 

\usepackage[ruled]{algorithm2e}
\usepackage{amssymb}
\usepackage{booktabs}

\SetAlFnt{\small}
\SetAlCapFnt{\small}
\SetAlCapNameFnt{\small}
\SetAlCapHSkip{0pt}
\IncMargin{-\parindent}

\doi{0000001.0000001}

\issn{1234-56789}

\begin{document}

\markboth{A. Joshi et al.}{Automatic Sarcasm Detection: A Survey}

\title{Automatic Sarcasm Detection: A Survey}
\author{ADITYA JOSHI
\affil{IITB-Monash Research Academy}
PUSHPAK BHATTACHARYYA
\affil{Indian Institute of Technology Bombay}
MARK J CARMAN
\affil{Monash University}
}

\begin{abstract}
Automatic sarcasm detection is the task of predicting sarcasm in text. This is a crucial step to sentiment analysis, considering prevalence and challenges of sarcasm in sentiment-bearing text. Beginning with an approach that used speech-based features, sarcasm detection has witnessed great interest from the sentiment analysis community. This paper is the first known compilation of past work in automatic sarcasm detection. We observe three milestones in the research so far: semi-supervised pattern extraction to identify implicit sentiment, use of hashtag-based supervision, and use of context beyond target text. In this paper, we describe datasets, approaches, trends and issues in sarcasm detection. We also discuss representative performance values, shared tasks and pointers to future work, as given in prior works. In terms of resources that could be useful for understanding state-of-the-art, the survey presents several useful illustrations - most prominently, a table that summarizes past papers along different dimensions such as features, annotation techniques, data forms, etc. 
\end{abstract}

%
%


%
%



\acmformat{Aditya Joshi, Pushpak Bhattacharyya and Mark James Carman. 2016. Automatic Sarcasm Detection: A Survey}

\begin{bottomstuff}
Author's addresses: Aditya Joshi, IITB-Monash Research Academy, IIT Bombay, Mumbai - 400 076.
\end{bottomstuff}

\maketitle

\textbf{This paper is an early draft of the survey that is being submitted to ACM CSUR. The stylesheet used in ACM Small, resulting in the footers, etc. that are seen in this draft. The paper has been uploaded to arXiv for feedback from stakeholders.}
\section{Introduction}

The Free Dictionary\footnote{\url{www.thefreedictionary.com}} defines sarcasm as a form of verbal irony that is intended to express contempt or ridicule\footnote{Sarcasm is a form of verbal irony. This explains the relationship between sarcasm and irony. Past work in sarcasm detection often says `we use the two interchangeably'}. The figurative nature of sarcasm makes it an often-quoted challenge for sentiment analysis~\cite{sabook}. It has an implied negative sentiment, but a positive surface sentiment. This led to interest in automatic sarcasm detection as a research problem. Automatic sarcasm detection refers to computational approaches to predict if a given text is sarcastic. This problem is hard because of nuanced ways in which sarcasm may be expressed. 

Starting with the earliest known work by ~\citeN{1} which deals with sarcasm detection in speech, the area has seen wide interest from the natural language processing community as well. Following that, sarcasm detection from text has extended to different data forms (tweets, reviews, TV series dialogues), and spanned several approaches (rule-based, supervised, semi-supervised). This synergy has resulted in interesting innovations for automatic sarcasm detection. The goal of this survey paper\footnote{\citeN{16} is a survey of linguistic challenges of computational irony. Their paper focuses on linguistic theories and possible applications of these theories for sarcasm detection. On the contrary, we deal with the computational angle, and present a survey of `\textit{computational}' sarcasm detection techniques.} is to look back at past work in computational sarcasm detection to enable new researchers to understand state-of-the-art.

Our paper looks at sarcasm detection in six steps: problem formulation, datasets, approaches, reported performance, trends and issues. We also discuss shared tasks related to sarcasm detection and future areas as pointed out in past work. 

The rest of the paper is organized as follows. Section~\ref{sec:linguistics} first describes sarcasm studies in linguistics. Section~\ref{sec:probdef} then presents different problem definitions for sarcasm detection. Sections~\ref{sec:datasets} and~\ref{sec:approaches} discuss datasets and approaches reported for sarcasm detection, respectively. Section~\ref{sec:trends} highlights trends underlying sarcasm detection, while Section ~\ref{sec:issues} discusses recurring issues. Section~\ref{sec:concl} concludes the paper.
\section{Sarcasm Studies in Linguistics}
\label{sec:linguistics}
Sarcasm as a linguistic phenomenon has been widely studied. Before we begin with approaches for automatic sarcasm detection, we present an introduction to sarcasm studies in linguistics.

Several representations and taxonomies for sarcasm have been proposed:
\begin{enumerate}
\item \citeN{campbell2012there} state that sarcasm occurs along several dimensions, namely, failed expectation, pragmatic insincerity, negative tension, and presence of a victim.  
\item \citeN{camp2012sarcasm} show that there are four types of sarcasm: (1) \textbf{Propositional}: Such sarcasm appears to be a non-sentiment proposition but has an implicit sentiment involved, (2) \textbf{Embedded}: This type of sarcasm has an embedded sentiment incongruity in the form of words and phrases themselves, (3) \textbf{Like-prefixed}: A like-phrase provides an implied denial of the argument being made, and (4) \textbf{Illocutionary}: This kind of sarcasm involves non-textual clues that indicate an attitude opposite to a sincere utterance. In such cases, prosodic variations play a role in sarcasm expression.
\item \textbf{6-tuple representation}: \citeN{ivanko2003context} define sarcasm as a 6-tuple consisting of $<$S, H, C, u, p, p'$>$ where:
\begin{center}
S = Speaker , H = Hearer/Listener \\
C = Context, u = Utterance \\
p = Literal Proposition \\
p' = Intended Proposition \\
\end{center}
The tuple can be read as `\textit{Speaker S generates an utterance u in Context C meaning proposition p but intending that hearer H understands p}'. Consider the following example. If a teacher says to a student, ``That's how assignments should be done!" and if the student knows that (s)he has barely completed the assignment, the student would understand the sarcasm. In context of the 6-tuple above, the properties of this sarcasm would be: \\
\textit{S: Teacher, H: Student\\
C: The student has not completed his/her assignment.\\
u: ``That's how assignments should be done!"\\
p: You have done a good job at the assignment.\\
p': You have done a bad job at the assignment.\\
}
\item \citeN{eisterhold2006reactions} state that sarcasm can be understood in terms of the response it elicits. They observe that the responses to sarcasm may be laughter, zero response, smile, sarcasm (in return), a change of topic (because the listener was not happy with the caustic sarcasm), literal reply and non-verbal reactions.
\item \textbf{Situational disparity theory}: According to \citeN{wilson2006pragmatics}, sarcasm arises when there is situational disparity between text and a contextual information. 
\item \textbf{Negation theory of sarcasm}: \citeN{giora1995irony} state that irony/sarcasm is a form of negation in which an explicit negation marker is lacking. In other words, when one expresses sarcasm, a negation is intended, without putting a negation word like `not'.
\end{enumerate}

In the context of the theories described here, some challenges typical to sarcasm are: (1) Identification of common knowledge, (2) Identification of what constitutes ridicule, (3) Speaker-listener context (\textit{i.e.}, knowledge shared by the speaker and the listener). As we will see in the next sections, the focus of automatic sarcasm detection approaches in the past has been (1) and (3) where they capture context using different techniques.
\section{Problem Definition}
\label{sec:probdef}
We now look at how the problem of automatic sarcasm detection has been defined, in past work. The most common formulation for sarcasm detection is a \textbf{classification} task. Given a piece of text, the goal is to predict whether or not it is sarcastic. However, past work varies in terms of what these output labels are. For example, understanding the relationship between sarcasm, irony and humor, \citeN{17} consider labels for the classifier as: politics, humor, irony and sarcasm. \citeN{13} use a similar formulation and provide pair-wise classification performance for these labels. 

Other formulations for sarcasm detection have also been reported. \citeN{41} deviate from the traditional classification definition and models sarcasm detection for dialogue as a sequence labeling task. Each utterance in a dialogue is considered to be an observed unit in this sequence, whereas sarcasm labels are the hidden variables whose values need to be predicted. \citeN{30} model sarcasm detection as a \textbf{sense disambiguation} task. They state that a word may have a literal sense and a sarcastic sense. Their goal is to identify the sense of a word in order to detect sarcasm. 

Table~\ref{tab:matrix} shows a matrix that summarizes past work in automatic sarcasm detection. While several interesting observations are possible from the table, two are key: (a) tweets are the predominant text form for sarcasm detection, and (b) incorporation of extra-textual context is a recent trend in sarcasm detection. 
\begin{table}
\centering
\tbl{Summary of sarcasm detection along different parameters\label{tab:matrix}}{
\begin{tabular}{|p{3.7cm}|p{0.15cm}|p{0.15cm}|p{0.15cm}|p{0.15cm}|p{0.15cm}|p{0.15cm}|p{0.15cm}|p{0.15cm}|p{0.15cm}|p{0.15cm}|p{0.15cm}|p{0.15cm}|p{0.15cm}|p{0.15cm}|p{0.15cm}|p{0.15cm}|p{0.15cm}|}
\hline
\textbf{}   & \multicolumn{3}{|c|}{\textbf{Datasets}} & \multicolumn{3}{|c|}{\textbf{Approach}} & \multicolumn{3}{|c|}{\textbf{Annotatn.}} & \multicolumn{5}{|c|}{\textbf{Features}} &  \multicolumn{3}{|c|}{\textbf{Context}} \\ \hline
\textbf{} &  \rotatebox[origin=c]{90}{\textbf{Short Text}} &  \rotatebox[origin=c]{90}{\textbf{Long Text}} &  \rotatebox[origin=c]{90}{\textbf{Other}} &  \rotatebox[origin=c]{90}{\textbf{Rule-based}} &  \rotatebox[origin=c]{90}{\textbf{Semi-superv.}} &  \rotatebox[origin=c]{90}{\textbf{Superv}} &  \rotatebox[origin=c]{90}{\textbf{Manual}} & \rotatebox[origin=c]{90}{\textbf{Distant}} &
\rotatebox[origin=c]{90}{\textbf{Other}} &
\rotatebox[origin=c]{90}{\textbf{Unigram}} & \rotatebox[origin=c]{90}{\textbf{Sentiment}} & \rotatebox[origin=c]{90}{\textbf{Pragmatic}} & \rotatebox[origin=c]{90}{\textbf{Patterns}}
& \rotatebox[origin=c]{90}{\textbf{Other}} & \rotatebox[origin=c]{90}{\textbf{Author}} & \rotatebox[origin=c]{90}{\textbf{Conversation}} & \rotatebox[origin=c]{90}{\textbf{Other}}    \\ \hline
\textbf{\cite{2}} & & & \checkmark & & & & \checkmark & & & \checkmark & & & & & & &  \\ \hline
\textbf{\cite{3}} & & \checkmark &  & &\checkmark & & \checkmark & & & \checkmark & & &\checkmark & & & &  \\ \hline
\textbf{\cite{4}} & & \checkmark &  & &\checkmark & & \checkmark & & & \checkmark & & &\checkmark & & & &  \\ \hline
\textbf{\cite{5}} & & & \checkmark & \checkmark & & & & & \checkmark & \checkmark & & & & & & & \\ \hline
\textbf{\cite{6}} & \checkmark & & & & &\checkmark & &\checkmark & & \checkmark &\checkmark &\checkmark & & & & & \\ \hline
\textbf{\cite{7}} & \checkmark & & & & & \checkmark & &\checkmark & & \checkmark&\checkmark &\checkmark & &\checkmark & & &  \\ \hline
\textbf{\cite{8}} & & \checkmark& & & &\checkmark & \checkmark & & &\checkmark & \checkmark& & &\checkmark & & & \\ \hline
\textbf{\cite{9}} & & \checkmark & & & & &\checkmark & & & & & & & & & & \\ \hline
\textbf{\cite{10}} & \checkmark &  & & &\checkmark & &\checkmark & & &\checkmark & &\checkmark &\checkmark & & & & \\ \hline
\textbf{\cite{11}} & &\checkmark & & &\checkmark & &\checkmark & & &\checkmark & & &\checkmark & & & &  \\ \hline
\textbf{\cite{12}} & \checkmark & & & & &\checkmark & &\checkmark & & &\checkmark & &\checkmark & &\checkmark & &  \\ \hline
\textbf{\cite{13}} & \checkmark & & & & &\checkmark & &\checkmark & &\checkmark &\checkmark &\checkmark & &\checkmark& & &  \\ \hline
\textbf{\cite{14}} &\checkmark & \checkmark & & & &\checkmark & &\checkmark & &\checkmark &\checkmark &\checkmark & &\checkmark& & &   \\ \hline
\textbf{\cite{15}} & & &\checkmark & & &\checkmark &\checkmark & & &\checkmark & & & &\checkmark & & &  \\ \hline
\textbf{\cite{17}} &\checkmark & & & & &\checkmark & &\checkmark & & &\checkmark & & &\checkmark & & &  \\ \hline
\textbf{\cite{18}} &\checkmark & & &\checkmark & & &\checkmark & & &\checkmark &\checkmark & & &\checkmark & & &  \\ \hline
\textbf{\cite{19}} & & \checkmark & & & & &\checkmark & & & & & & & & & &  \\ \hline
\textbf{\cite{20}} & &\checkmark & & & &\checkmark & & &\checkmark &\checkmark &\checkmark &\checkmark & & &\checkmark & &  \\ \hline
\textbf{\cite{21}} & \checkmark & & & & &\checkmark & & &\checkmark &\checkmark & & &\checkmark & & & &  \\ \hline
\textbf{\cite{22}} & \checkmark &\checkmark & & & &\checkmark &\checkmark &\checkmark & &\checkmark &\checkmark &\checkmark &\checkmark & & & & \\ \hline
\textbf{\cite{23}} &\checkmark & & &\checkmark & & &\checkmark & & &\checkmark &\checkmark & & & &\checkmark & &  \\ \hline
\textbf{\cite{24}} &\checkmark & & & & &\checkmark & &\checkmark & & \checkmark & \checkmark& & &\checkmark & \checkmark&\checkmark &  \\ \hline
\textbf{\cite{25}} &\checkmark & & & & &\checkmark & &\checkmark & &\checkmark &\checkmark &\checkmark & &\checkmark &\checkmark &\checkmark & \checkmark  \\ \hline
\textbf{\cite{26}} & &\checkmark & & & &\checkmark &\checkmark & & &\checkmark &\checkmark & & &\checkmark  &  &\checkmark & \checkmark \\ \hline
\textbf{\cite{27}} &\checkmark  & & &\checkmark  &\checkmark  &\checkmark  & &\checkmark  & &\checkmark  & &\checkmark  & & & & &  \\ \hline
\textbf{\cite{28}} &\checkmark  & & & & &\checkmark & &\checkmark & &\checkmark &\checkmark &\checkmark & &\checkmark & & &  \\ \hline
\textbf{\cite{29}} &\checkmark & & & & &\checkmark & & &\checkmark & \checkmark & & & & & &\checkmark &  \\ \hline
\textbf{\cite{30}} & & &\checkmark & & & &\checkmark & & & \checkmark& & & &\checkmark & & & \\ \hline
\textbf{\cite{33}} &\checkmark &\checkmark & & & &\checkmark & &\checkmark & &\checkmark &\checkmark &\checkmark & &\checkmark & & & \\ \hline
\textbf{\cite{32}} &\checkmark & & & &\checkmark & & &\checkmark & &\checkmark &\checkmark & &\checkmark & & & & \\ \hline
\textbf{\cite{34}} &\checkmark & & & & &\checkmark & &\checkmark& &\checkmark& &\checkmark& &\checkmark & & & \\ \hline
\textbf{\cite{36}} &\checkmark & & & & &\checkmark &\checkmark& & &\checkmark & &\checkmark & &\checkmark & & & \\ \hline
\textbf{\cite{35}} & \checkmark & & & & &\checkmark & &\checkmark & & \checkmark &\checkmark &\checkmark & & & & & \\ \hline
\textbf{\cite{40}} & \checkmark &\checkmark & \checkmark &\checkmark & & &\checkmark & & &\checkmark  &\checkmark &\checkmark & &\checkmark& & & \\ \hline
\textbf{\cite{41}} &  & & \checkmark & & & \checkmark &\checkmark & & &\checkmark  &\checkmark & & & &\checkmark &\checkmark & \\\hline
\textbf{\cite{42}} & \checkmark & & & & &\checkmark &\checkmark& & &\checkmark  & & & & &\checkmark & \checkmark & \checkmark\\ \hline
\textbf{\cite{43}} &\checkmark  & & & & &\checkmark & & \checkmark& &  & & & & &\checkmark & & \\ \hline
\textbf{\cite{44}} & \checkmark & & & & &\checkmark &\checkmark & & &  & & & & & & & \\\hline
\textbf{\cite{45}} &\checkmark  & & & &\checkmark & & &\checkmark& &\checkmark & &\checkmark &\checkmark & & & & \\\hline
\textbf{\cite{47}} &  & &\checkmark & & &\checkmark & & \checkmark& &\checkmark & & & &\checkmark & & & \\ \hline
\end{tabular}}
\end{table}

\textbf{A note on languages}\\
Most research in sarcasm detection exists for English. However, some research in the following languages has also been reported: Chinese~\cite{33}, Italian~\cite{21}, Czech~\cite{37}, Dutch~\cite{12}, Greek~\cite{38}, Indonesian~\cite{39} and Hindi~\cite{46}.

\begin{table}[h!]
\centering
\tbl{Summary of sarcasm-labeled datasets\label{tab:datasets}}{
\begin{tabular}{|l|p{10cm}|}
\hline
\textbf{Text form}                                                    & \textbf{Related Work}                                                                                                                           \\ \hline
Tweets                                                                & \textbf{Manual}: \cite{10,18,37,40,42}   \\\hline
 & \textbf{Hashtag-based}: \cite{4,6,7,13,21,22,27,32,12,36,29,17,25,34,23,24,42} \\ \hline
Reddits                                                               & \cite{19,26}               \\ \hline
Long text (Reviews, etc.) & \cite{11,14,8,20,33,9}                                                                                                                        \\ \hline
Other datasets                                                        & \cite{1,2,5,15,30,41,42}                                                                                                                            \\ \hline
\end{tabular}}
\end{table}
\section{Datasets}
\label{sec:datasets}
This section describes different datasets used for experiments in sarcasm detection. We divide them into three classes: short text (typically characterized by noise and situations where length is limited by the platform, as in tweets), long text (such as discussion forum posts) and other datasets.
\subsection{Short text}
Social media makes available several forms of data. However, because of word limit, text on some platforms tends to be short. However, datasets of tweets have been popular for sarcasm detection. This may be because of availability of the Twitter API and popularity of twitter as a medium. One approach to obtain labels for tweets is manual annotation. \citeN{10} introduce a dataset of tweets, manually annotated as sarcastic or not. \citeN{18} study sarcastic tweets and their impact to sarcasm classification. They experiment with around 600 tweets which are marked for subjectivity, sentiment and sarcasm. \citeN{37} present a dataset of 7000 manually labeled tweets in Czech.

The second technique to create datasets is the use of hashtag-based supervision. Many approaches use hashtags in tweets as indicators of sarcasm, to create labeled datasets. The popularity of this approach (over manual annotation) can be attributed to various factors: (a) No one but the author of a tweet can determine if it was sarcastic. A hashtag is a label provided by authors themselves, (b) The approach allows creation of large-scale datasets. In order to create such a dataset, tweets containing particular hashtags are labeled as sarcastic. \citeN{4} use a dataset of tweets, which are labeled with hashtags such as \#sarcasm, \#sarcastic, \#not, etc. \citeN{6} also use hashtag-based supervision for tweets. However, they retain examples where it occurs at the end of a tweet but eliminate cases where the hashtag is a part of the running text. For example, `\#sarcasm is popular among teens' is eliminated. \citeN{7} use similar approach. \citeN{13} use a dataset of 40000 tweets labeled as sarcastic or not, using hashtags. \citeN{27} present hashtag-annotated dataset of tweets: 1000 trial, 4000 development and 8000 test tweets.  \citeN{12} use`\#not' to download and label their tweets. \citeN{17} create a dataset using hashtag-based supervision based on hashtags indicated by multiple labels: politics, sarcasm, humor and irony. Other works using this approach have also been reported~\cite{21,22,32,36,42}.

However, use of distant supervision using hashtags poses challenges, and may require quality control. To ensure quality, \citeN{25} label tweets as: the positive tweets are the ones containing \#sarcasm – the negative tweets are assumed to be the one not containing these labels. \citeN{34} present a dataset of 8K tweets where the initial label is based on the hashtag. To ensure quality, these tweets are additionally labelled by annotators. 

Twitter also provides access to additional context. Hence, in order to predict sarcasm, supplementary datasets\footnote{`Supplementary' datasets refer to text that does not need to be annotated but that will contribute to the judgment of the sarcasm detector} have also been used for sarcasm detection.  \citeN{23} use a supplementary set of complete twitter timeline (limited to 3200 tweets, by Twitter) to establish context for a given dataset of tweets. \cite{24} use a dataset of tweets, labeled by hashtag-based supervision – along with a historical context of 80 tweets per author. 

Like supplementary datasets, supplementary annotation (\textit{i.e.}, annotation apart from sarcasm/non-sarcasm) has also been explored. \citeN{40} capture cognitive features based on eye-tracking. They employ annotators who are asked to determine the sentiment (and not `sarcasm/not-sarcasm', since, as per their claim, it can result in priming) of a text. While the annotators read the text, their eye movements are recorded by an eye-tracker. This eye-tracking information serves as supplementary annotation.

Other social media text includes reddits. \citeN{19} create a corpus of reddit posts of 10K sentences, from 6 reddit topics.~\cite{26} present a dataset of reddit comments - 5625 sentences.
\subsection{Long text}
Reviews and discussion forum posts have also been used as sarcasm-labeled datasets. \citeN{11} present Internet Argument Corpus that marks a dataset of discussion forum posts with multiple labels – one of them being sarcasm. \citeN{14} create a dataset of movie reviews, book reviews and news articles marked with sarcasm and sentiment. \citeN{8} deal with products that saw a spate of sarcastic reviews all of a sudden. The dataset consists of 11000 reviews. \citeN{9} use a sarcasm-labeled dataset of around 1000 reviews. \citeN{20} create a labeled set of 1254 Amazon reviews, out of which 437 are ironic. \citeN{3} consider a large dataset of 66000 amazon reviews. \citeN{33} use a dataset from multiple sources such as Amazon, Twitter, Netease and Netcena. In these cases, the datasets are manually annotated because markers like hashtags are not available.

\subsection{Other datasets}
Other novel datasets have also been used. \citeN{1} use 131 call center transcripts. Each occurrence of `yeah right' is marked as sarcastic or not. The goal is to identify which `yeah right' is sarcastic. \citeN{2} use 20 sarcastic excerpts and 15 non-sarcastic excerpts, which are marked by 101 students. The goal is to identify lexical indicators of sarcasm. \citeN{5} focus on identifying which similes are sarcastic. Hence, they first search the web for the pattern `* as a *'. This results in 20,000 distinct similes which are then marked as sarcastic or not. \citeN{15} create a crowdsourced dataset of sentences from a MTV show, Daria. On similar lines, \citeN{41} report their results on a manually annotated dataset of the TV Series `Friends'. Every `utterance' (sic) in a scene is annotated with two labels: sarcastic or not sarcastic. \citeN{30} use a crowdsourcing tool to obtain a non-sarcastic version of a sentence if applicable. For example `Who doesn't love being ignored' is expected to be corrected to `Not many love being ignored'. \citeN{40} create a manually labeled dataset of quotes from a website called sarcasmsociety.com.

\section{Approaches}
\label{sec:approaches}
Following the discussion on datasets, we now describe approaches used for sarcasm detection.
 In general, approaches to sarcasm detection can be classified into: rule-based, statistical and deep learning-based approaches. We look at these approaches in the next subsections. Following that, we describe shared tasks in conferences that deal with sarcasm detection.

\subsection{Rule-based Approaches}
\label{sec:ruleb}
Rule-based approaches attempt to identify sarcasm through specific evidences. These evidences are captured in terms of rules that rely on indicators of sarcasm. \citeN{5} focus on identifying whether a given simile (of the form `\textit{* as a *}') is intended to be sarcastic. They use Google search in order to determine how likely a simile is. They present a 9-step approach where at each step/rule, a simile is validated using the number of search results. A strength of this approach is that they present an error analysis corresponding to multiple rules. \citeN{18} propose that hashtag sentiment is a key indicator of sarcasm. Hashtags are often used by tweet authors to highlight  sarcasm, and hence, if the sentiment expressed by a hashtag does not agree with rest of the tweet, the tweet is predicted as sarcastic. They use a hashtag tokenizer to split hashtags made of concatenated words. \citeN{32} present two rule-based classifiers. The first uses a parse--based lexicon generation algorithm that creates parse trees of sentences and identifies situation phrases that bear sentiment. If a negative phrase occurs in a positive sentence, it is predicted as sarcastic. The second algorithm aims to capture hyperboles by using interjection and intensifiers occur together. \citeN{10} present rule-based classifiers that look for a positive verb and a negative situation phrase in a sentence. The set of negative situation phrases are extracted using a well-structured, iterative algorithm that begins with a bootstrapped set of positive verbs and iteratively expands both the sets (positive verbs and negative situation phrases). They experiment with different configurations of rules such as restricting the order of the verb and situation phrase.

\subsection{Statistical Approaches}
\label{sec:stat}
Statistical approaches to sarcasm detection vary in terms of features and learning algorithms. We look at the two in forthcoming subsections.

\subsubsection{Features Used}

In this subsection, we look at the set of features that have been reported for statistical sarcasm detection. Most approaches use bag-of-words as features. However, in addition to these, there are peculiar features introduced in different works. Table~\ref{tab:features} summarizes sets of features used for statistical approaches. In this subsection, we focus on features related to the text to be classified. Contextual features (\textit{i.e.}, features that use information beyond the text to be classified) are described in a latter subsection.

\citeN{3} design pattern-based features that indicate presence of discriminative patterns as extracted from a large sarcasm-labeled corpus. To allow generalized patterns to be spotted by the classifiers, these pattern-based features take real values based on three situations: exact match, partial overlap and no match. \citeN{6} use sentiment lexicon-based features. In addition, pragmatic features like emoticons and user mentions are also used. \citeN{7} introduce features related to ambiguity, unexpectedness, emotional scenario, etc. Ambiguity features cover structural, morpho-syntactic, semantic ambiguity, while unexpectedness features measure semantic relatedness. \citeN{10} use a set of patterns, specifically positive verbs and negative situation phrases, as features for a classifier (in addition to a rule-based classifier). \citeN{12} introduce bigrams and trigrams as features. \citeN{13} explore skip-gram and character n-gram-based features. \citeN{18} include seven sets of features. Some of these are maximum/minimum/gap of intensity of adjectives and adverbs, max/min/average number of synonyms and synsets for words in the target text, etc. Apart from a subset of these, \citeN{21} use frequency and rarity of words as indicators. \citeN{20} incorporate ellipsis, hyperbole and imbalance in their set of features. \citeN{22} use features corresponding to the linguistic theory of incongruity. The features are classified into two sets: implicit and explicit incongruity-based features. \citeN{37} use word-shape and pointedness features given in the form of 24 classes. \citeN{24} use extensions of words, number of flips, readability features in addition to others. \citeN{28} present features that measure semantic relatedness between words using Wordnet-based similarity. \citeN{33} introduce POS sequences and semantic imbalance as features. Since they also experiment with Chinese datasets, they use language-typical features like use of homophony, use of honorifics, etc. \citeN{40} conduct additional experiments with human annotators where they record their eye movements. Based on these eye movements, they design a set of gaze based features such as average fixation duration, regression count, skip count, etc. In addition, they also use complex gaze-based features based on saliency graphs which connect words in a sentence with edges representing saccade between the words.

\subsubsection{Learning Algorithms}
\label{sec:learn}
A variety of classifiers have been experimented for sarcasm detection. Most work in sarcasm detection relies on SVM~\cite{22,1,2,3,4} (or SVM-Perf as in the case of \citeN{47}). \citeN{6} use SVM with SMO and logistic regression. Chi-squared test is used to identify discriminating features. \citeN{8} use Naive Bayes and SVM. They also show Jaccard similarity between labels and the features. \citeN{10} compare rule-based techniques with a SVM-based classifier. \citeN{12} use balanced winnow algorithm in order to determine high-ranking features. \citeN{13} use Naive Bayes and decision trees for multiple pairs of labels among irony, humor, politics and education. \citeN{25} use binary logistic regression. \citeN{29} use SVM-HMM in order to incorporate sequence nature of output labels in a conversation. \citeN{33} compare several classification approaches including bagging, boosting, etc. and show results on five datasets. On the contrary, \citeN{41} experimentally validate that for conversational data, sequence labeling algorithms perform better than classification algorithms. They use SVM-HMM and SEARN as the sequence labeling algorithms. 
\subsection{Deep Learning-based Approaches}
\label{sec:deeplb}
As architectures based on \textbf{deep learning} techniques gain popularity, few such approaches have been reported for automatic sarcasm detection as well. \citeN{47} use similarity between word embeddings as features for sarcasm detection. They augment features based on similarity of word embeddings related to most congruent and incongruent word pairs, and report an improvement in performance. The augmentation is key because they observe that using these features alone does not suffice. \citeN{43} present a novel convolutional network-based that learns user embeddings in addition to utterance-based embeddings. The authors state that it allows them to learn user-specific context. They report an improvement of 2\% in performance. \citeN{44} use a combination of convolutional neural network, LSTM followed by a DNN. They compare their approach against recursive SVM, and show an improvement in case of deep learning architecture.
\begin{table}[ht!]
\centering
\tbl{Summary of Features used for Statistical Classifiers\label{tab:features}}{
\begin{tabular}{|p{4cm}|p{8cm}|}
\hline
\textbf{}   & \textbf{Salient Features}                                                                           \\ \hline
\cite{3}  & Sarcastic patterns, Punctuations                                                                    \\ 
\cite{6}  & User mentions, emoticons, unigrams, sentiment-lexicon-based features                                \\ 
\cite{7}  & Ambiguity-based, semantic relatedness                                                               \\ 
\cite{8}  & N-grams, POS N-grams                                                                                \\ 
\cite{10} & Sarcastic patterns (Positive verbs, negative phrases)                                               \\ 
\cite{12} & N-grams, emotion marks, intensifiers                                                                \\ 
\cite{13} & Skip-grams, Polarity skip-grams                                                                     \\ 
\cite{17} & Synonyms, Ambiguity, Written-spoken gap                                                             \\ 
\cite{20} & Interjection, ellipsis, hyperbole, imbalance-based                                                  \\ 
\cite{21} & Freq. of rarest words, max/min/avg \# synsets, max/min/avg \# synonyms                              \\ 
\cite{22} & Unigrams, Implicit incongruity-based, Explicit incongruity-based                                    \\ 
\cite{24} & Readability, flips, etc.                                                                            \\ 
\cite{28} & Length, capitalization, semantic similarity                                                         \\ 
\cite{33} & POS sequences, Semantic imbalance. Chinese-specific features such as homophones, use of honorifics \\ 
\cite{37} & Word shape, Pointedness, etc. \\ 
\cite{40} & Cognitive features derived from eye-tracking experiments \\
\cite{45} & Pattern-based features along with word-based, syntactic, punctuation-based and sentiment-related features \\
\cite{47} & Features based on word embedding similarity \\ 
\hline
\end{tabular}}
\end{table}
\subsection{Shared Tasks}
\label{sec:sharedt}
Shared tasks in conferences allow a common dataset to be shared across multiple teams, for a comparative evaluation. Two shared tasks related to sarcasm detection have been conducted in the past. \citeN{27} is a shared task from SemEval-2015 that deals with sentiment analysis of figurative language. The organizers provided a dataset of ironic and metaphorical statements labeled as positive, negative and neutral. The participants were expected to correctly identify the sentiment polarity in case of figurative expressions like irony. The teams that participated in the shared task used affective resources, character n-grams, etc. The winning team used ``four lexica, one that was
automatically generated and three than were
manually crafted. (sic)''. The second shared task was a data science contest organized as a part of PAKDD 2016~\footnote{http://www.parrotanalytics.com/pacific-asia-knowledge-discovery-and-data-mining-conference-2016-contest/}. The dataset provided consists of reddit comments labeled as either sarcastic or non-sarcastic.
\section{Reported Performance}
\label{sec:reportedv}
Table~\ref{tab:results} summarizes reported values from past works. The values may not be directly comparable because they work with different kinds of datasets, and report different metrics. However, the table does provide a ballpark estimate of performance of sarcasm detection. \citeN{6} show that unigram-based features outperform the use of a subset of words as derived from a sentiment lexicon. They compare the accuracy of the sarcasm classifier with the human ability to detect sarcasm. While the best classifier achieves 57.41\%, the human performance for sarcasm identification is 62.59\%. \citeN{8} observe that sentiment-based features are their top discriminating features. The logistic classifier in \citeN{15} results in an accuracy of 81.5\%. \citeN{22} present an analysis of errors like incongruity due to numbers and granularity of annotation. \citeN{24} show that historical features along with flip-based features are the most discriminating features, and result in an accuracy of 83.46\%. These are also the features presented in a rule-based setting by \cite{23}. 
\begin{table}[]
\tbl{Summary of Performance Values; Precision/Recall/F-measures and Accuracy values are indicated in percentages\label{tab:results}}{
\begin{tabular}{|l|l|l|}
\hline
           & Details                  & Reported Performance \\ \hline
\cite{1} & Conversation transcripts & F: 70, Acc: 87   \\ \hline
\cite{4} & Tweets                   & F: 54.5 Acc: 89.6  \\   \hline  
\cite{6} & Tweets & A: 75.89 \\ \hline
\cite{7} & Irony vs  general & A: 70.12, F: 65 \\ \hline
\cite{8} & Reviews & F: 89.1, P: 88.3, R: 89.9 \\ \hline
\cite{10} & Tweets & F: 51, P: 44, R: 62 \\ \hline
\cite{11} & Discussion forum posts & F: 69, P: 75, R: 62 \\ \hline
\cite{12} & Tweets & AUC: 0.76 \\ \hline
\cite{13} & Irony vs humor & F: 76 \\ \hline
\cite{15} & Speech data & Acc: 81.57 \\ \hline
\cite{35} & Reviews & F: 75.7 \\ \hline
\cite{47} & Book snippets & F: 80.47 \\ \hline
\cite{24} & Tweets & Acc: 83.46, AUC: 0.83 \\ \hline
\cite{25} & Tweets & Acc: 85.1 \\ \hline
\cite{27} & Tweets & Cosine: 0.758, MSE: 2.117 \\ \hline
\cite{34} & Tweets & F: 83.59, Acc: 94.17 \\ \hline
\cite{22} & Tweets/Disc. Posts & F: 88.76/64 \\ \hline
\cite{23} & Tweets & F: 88.2 \\ \hline
\cite{29} & Tweets & Macro-F: 69.13 \\ \hline
\cite{41} & TV transcripts & F: 84.4 \\ \hline
\cite{42} & Tweets & AUC: 0.6 \\ \hline
\cite{20} & Reviews & F: 71.3 \\ \hline
\cite{28} & Irony vs politics & F: 81 \\ \hline
\end{tabular}}
\end{table}
\section{Trends in Sarcasm Detection}
\label{sec:trends}
\begin{figure*}[ht!]
\centering
  \includegraphics[width=0.9\linewidth]{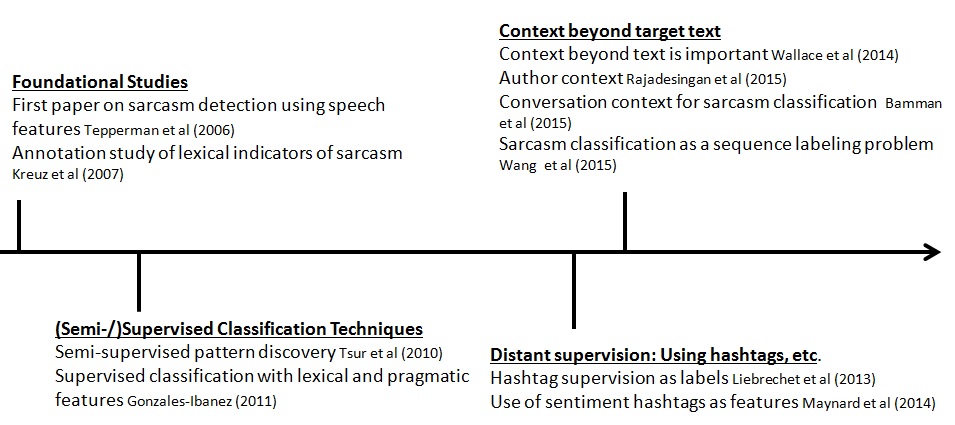}
  \caption{Trends in Sarcasm Detection Research}
  \label{fig:boat1}
\end{figure*}
In the previous sections, we looked at the datasets, approaches and performance values of past work in sarcasm detection. In this section, we delve into trends observed in sarcasm detection research. Figure~\ref{fig:boat1} summarizes these trends. Representative work in each area are indicated in the figure. As seen in the figure, there have been four key milestones. Following fundamental studies, supervised/semi-supervised sarcasm classification approaches were explored. These approaches focused on using specific patterns or novel features. Then, as twitter emerged as a viable source of data, hashtag-based supervision became popular. Recently, using context beyond the text to be classified has become popular.

In the rest of this section, we describe in detail two of these trends: (a) discovery of sarcastic patterns, and use of these patterns as features, and (b) use of contextual information \textit{i.e.}, information beyond the target text for sarcasm detection. We describe the two trends in detail in the forthcoming subsections.

\subsection{Pattern discovery}
Discovering sarcastic patterns was an early trend in sarcasm detection. Several approaches dealt with extracting patterns that are indicative of sarcasm, or carry implied sentiment. These patterns may then be used as features for a statistical classifier, or as rules in a rule-based classifier.
\citeN{3} extract sarcastic patterns from a seed set of labeled sentences. They first select words that either occur more than an upper threshold or less than a lower threshold. Among these words, identify a large set of candidate patterns. The patterns which occur discriminatively in either classes are then selected. \citeN{37,45} also use a similar approach for Czech and English tweets. 

\citeN{10} hypothesize that sarcasm occurs due to a contrast between positive verbs and negative situation phrases. To discover a lexicon of these verbs and phrases, they propose an iterative algorithm. Starting with a seed set of positive verbs, they identify discriminative situation phrases that occur with these verbs in sarcastic tweets. These phrases are then used to identify other verbs. The algorithm iteratively appends to the list of known verbs and phrases. \citeN{22} adapt this algorithm by eliminating subsumption, and show that it adds value. \citeN{11} begin with a seed set of nastiness and sarcasm patterns, created using Amazon Mechanical Turk. They train a high precision sarcastic post classifier, followed by a high precision non-sarcastic post classifier. These two classifiers are then used to generate a large labeled dataset from a bootstrapped set of patterns.

\subsection{Role of context in sarcasm detection}
A recent trend in sarcasm detection is the use of context. The term context here refers to any information beyond the text to be predicted, and beyond common knowledge. In the rest of this section, we refer to the textual unit to be classified as `target text'. As we will see, this context may be incorporated in a variety of ways - in general, using supplementary data or using supplementary information from the source platform of the data. \citeN{19} describe an annotation study that first highlighted the need of context for sarcasm detection. The annotators mark reddit comments with sarcasm labels. During this annotation, annotators often request for additional context in the form of reddit comments. The authors also present a transition matrix that shows how many times authors change their labels after the context is displayed to them.

Following this observation and the promise of context for sarcasm detection, several recent approaches have looked at ways of incorporating it. The contexts that have been reported are of three types:
\begin{enumerate}\setlength \itemsep{0cm}
\item \textbf{Author-specific context} refers to textual footprint of the author of the target text.  For example, \citeN{23} follow the intuition that `A tweet is sarcastic either because it has words of contrasting sentiment in it, or because there is sentiment that contrasts with the author's historical sentiment'. Historical tweets by the same author are considered as the context. Named entity phrases in the target tweet are looked up in the timeline of the author in order to gather the true sentiment of the author. This historical sentiment is then used to predict whether the author is likely to be sarcastic, given the sentiment expressed towards the entity in the target tweet. \citeN{24} incorporate context about author using the author's past tweets. This context is captured as features for a classifier. The features deal with various dimensions. They use features about author's familiarity with twitter (in terms of use of hashtags), familiarity with language (in terms of words and structures), and familiarity with sarcasm. \citeN{25} consider author context in features such as historical salient terms, historical topic, profile info, historical sentiment (how likely is he/she to be negative), etc. \citeN{43} capture author-specific embeddings for a neural network based architecture.
\item \textbf{Conversation context} refers to text in the conversation of which the target text is a part. This incorporates the discourse structure of a conversation. \citeN{25} capture conversational context using pair-wise Brown features between the previous tweet and the target tweet. In addition, they also use `audience' features. These are author features of the tweet author who responded to the target tweet. \citeN{22} show that concatenation of the previous post in a discussion forum thread along with the target post leads to an improvement in precision. \citeN{26} look at comments in the thread structure to obtain context for sarcasm detection. To do so, they use the subreddit name, and noun phrases from the thread to which the target post belongs. \citeN{29} use \textbf{sequence labeling} technique to capture this context. For a sequence of tweets in a conversation, they estimate the most probable sequence of three labels: happy, sad and sarcastic, for the last tweet in the sequence. A similar approach is used in ~\cite{41} for sarcastic/non-sarcastic labels. 
\item \textbf{Topical context}: This context follows the intuition that some topics are likely to evoke sarcasm more commonly than others. \citeN{29} also use topical context. To predict sarcasm in a tweet, they download tweets containing a hashtag in the tweet. Then, based on timestamps, they create a sequence of these tweets and again use sequence labeling to detect sarcasm in the target tweet (the last in the sequence).
\end{enumerate}
\section{Issues in Sarcasm Detection}
\label{sec:issues}
The current set of techniques in sarcasm detection also results in recurring issues that are handled in different ways by different prior works. In this section, we focus on three important issues. The first set of issues deal with \textbf{data}: hashtag-based supervision, data imbalance and inter-annotator agreements. The second issue deals with a specific kind of \textbf{features} that have been used for classification: sentiment as a label. Finally, the third issue lies in the context of \textbf{classification} techniques where we look at how past works handle dataset skews.
\subsection{Issues with Data}
Although hashtag-based labeling can provide large-scale supervision, the quality of the dataset may become doubtful. This is particularly true in case of use of \#not to indicate insincere sentiment. \citeN{12} show how \#not can be used to express sarcasm - while the rest of the sentence is non-sarcastic. For example, `\textit{I totally love bland food. \#not}'. The speaker expresses sarcasm through \#not. In most reported works that use hashtag-based supervision, the hashtag is removed in the pre-processing step. This reduces the sentence above to '\textit{I love bland food}' - which may not have a sarcastic interpretation, unless author's context is incorporated. To mitigate this problem, a new trend is to validate on multiple datasets - some annotated manually while others annotated through hashtags~\cite{22,44,45}. \citeN{44} train their deep learning-based model using a large dataset of hashtag-annotated tweets, but use a test set of manually annotated tweets. 

In addition, since sarcasm is a subjective phenomenon, the inter-annotator agreement values reported in past work are diverse. \citeN{3} indicate an agreement of 0.34. The value in case of \citeN{1} is 52.73\%, in case of \citeN{34} is 0.79 while for \citeN{10}, it is 0.81. \citeN{joshicultures} perform an interesting study on cross-cultural sarcasm annotation. They compare annotations by Indian and American annotators, and show that Indian annotators agree with each other more than their American counterparts. They also give examples to elicit these differences. For example, `It's sunny outside and I am at work. Yay' is considered sarcastic by the American annotators, but non-sarcastic by Indian annotators due to typical Indian climate.

\subsection{Issues with features: Sentiment as feature}
One question that many papers deliberate is if sentiment can be used as a feature for sarcasm detection. The motivation behind sarcasm detection is often pointed as sarcastic sentences misleading a sentiment classifier. However, several approaches use sentiment as an input to the sarcasm classifier. It must, however, be noted that these approaches require `surface polarity' – the apparent polarity of a sentence.  \citeN{32} describe a rule-based approach that predicts a sentence as sarcastic if a negative phrase occurs in a positive sentence. As described earlier, \citeN{23} use sentiment of a past tweet by the author to predict sarcasm. In a statistical classifier, surface polarity may be used directly as a feature use polarity of the tweet as a feature~\cite{7,22,24,25}. \citeN{13} capture polarity in terms of two emotion dimensions: activation and pleasantness. \citeN{20} incorporate sentiment imbalance as a feature. Sentiment imbalance is a situation where star rating of a review disagrees with the surface polarity. \citeN{36} cascade sarcasm detection and sentiment detection, and observe an improvement of 4\% in accuracy when sentiment detection is aware of sarcastic nature.
\subsection{Dealing with Dataset Skews}
Sarcasm is an infrequent phenomenon of sentiment expression. This skew also reflects in datasets. \citeN{3} use a dataset with a small set of sentences are marked as sarcastic. 12.5\% of tweets in the Italian dataset given by \citeN{21} are sarcastic.  On the other hand, \citeN{15} present a balanced dataset of 15k tweets. \citeN{12} state that ``detecting sarcasm is like a needle in a haystack". In some papers, the technique used is designed to work around existing skew. \citeN{33} present a multi-strategy ensemble learning approach is used that uses ensembles and majority voting. \citeN{47} use SVM-perf that performs F-score optimization. Similarly, in order to deal with sparse features and skew of data, \citeN{26}  introduce a LSS-regularization strategy. Thus, they use a sparsifying L1 regularizer over contextual features and L2-norm for bag of word features. Since AUC is known to be a better indicator than F-score for skewed data, \citeN{12} report AUC for balanced as well as skewed datasets, to demonstrate the benefit of their classifier. Another methodology to ascertain benefit of a given approach withstanding data skew is by \citeN{42}. They compare performance of sarcasm classification across two dimensions: type of annotation (manual versus hashtag-supervised) and data skew.

\section{Conclusion \& Future Directions}
\label{sec:concl}
Sarcasm detection research has grown significantly in the past few years, necessitating a look-back at the overall picture that these individual works have led to. This paper surveys approaches for automatic sarcasm detection. We observed three milestones in the history of sarcasm detection research: semi-supervised pattern extraction to identify implicit sentiment, use of hashtag-based supervision, and use of context beyond target text. We tabulated datasets and approaches that have been reported. Rule-based approaches capture evidences of sarcasm in the form of rules such as sentiment of hashtag not matching sentiment of rest of the tweet. Statistical approaches use features like sentiment changes. To incorporate context, additional features specific to the author, the conversation and the topic have been explored in the past. We also highlight three issues in sarcasm detection: the relationship between sentiment and sarcasm, and data skew in case of sarcasm-labeled datasets. Our table that compares all past papers along dimensions such as approach, annotation approach, features, etc. will be useful to understand the current state-of-art in sarcasm detection research.

 Based on our survey of these works, we propose following possible directions for future:
\begin{enumerate}\setlength \itemsep{0cm}
\item \textbf{Implicit sentiment detection \& sarcasm}: Based on past work, it is well-established that sarcasm is closely linked to sentiment incongruity~\cite{22}. Several related works exist for detection of implicit sentiment in sentences, as in the case of `\textit{The phone gets \underline{heated} quickly}' v/s `\textit{The induction cooktop gets \underline{heated} quickly}'. This will help sarcasm detection, following the line of semi-supervised pattern discovery.
\item \textbf{Incongruity in numbers}: \citeN{22} point out how numerical values convey sentiment and hence, is related to sarcasm. Consider the example of `\textit{Took 6 hours to reach work today. \#yay}'. This sentence is sarcastic, as opposed to `\textit{Took 10 minutes to reach work today. \#yay}'.
\item \textbf{Coverage of different forms of sarcasm}: In Section 2, we described four species of sarcasm: propositional, lexical, like-prefixed and illocutionary sarcasm. We observe that current approaches are limited in handling the last two forms of sarcasm: like-prefixed and illocutionary. Future work may focus on these forms of sarcasm.
\item \textbf{Culture-specific aspects of sarcasm detection}: As shown in \citeN{33}, sarcasm is closely related to language/culture-specific traits. Future approaches to sarcasm detection in new languages will benefit from understanding such traits, and incorporating them into their classification frameworks. \citeN{joshicultures} show that American and Indian annotators may have substantial disagreement in their sarcasm annotations - however, this sees a non-significant degradation in the performance of sarcasm detection.
\item \textbf{Deep learning-based architectures}: Very few approaches have explored deep learning-based architectures so far. Future work that uses these architecture may show promise.
\end{enumerate}

\bibliographystyle{acm}
\bibliography{citation}
\end{document}